\documentclass{l4dc2027}

\title[Screened HardNet++]{
Scaling Up Nonlinear Constraint Enforcement in Neural Networks via \\ Adaptive Screening
}
\usepackage{times}
\usepackage{wrapfig}
\usepackage{booktabs}

\author{%
 \Name{Andrea Goertzen\nametag{\thanks{These authors contributed equally.}}} \Email{agoertz@mit.edu}\\
 \Name{Juho Bae\nametag{\footnotemark[1]}} \Email{juhobae@mit.edu}\\
\Name{Navid Azizan} \Email{azizan@mit.edu}\\
 \addr Massachusetts Institute of Technology%
}
\newcommand{\name}{HardNet++} 

\begin{document}

\maketitle

\thispagestyle{empty} 

\begin{abstract}%

 Hard-constrained neural networks can provide fast surrogates for optimization and simulation problems, but iterative constraint-enforcement layers become costly when the number of constraints is large. To address this gap, we introduce \emph{Screened \name}, a nonlinear constraint-enforcement method that iteratively corrects neural-network outputs using local constraint linearizations while adaptively selecting active constraints. At each iteration, the proposed method uses scaled margins to screen out inequalities that are safely within their bounds, thereby avoiding computation involving constraints that are safely inactive. We show that this screening rule retains the exponential decay of constraint violations of HardNet++. Furthermore, we establish universal approximation for any fixed number of iterations and, under stronger assumptions, for the limiting correction. We also show that raw-network approximation-error bounds transfer uniformly over iteration counts and batch sizes, so a sufficient raw-network parameter budget for a prescribed accuracy can be chosen independently of the correction depth. We evaluate the approach on AC optimal power flow for the IEEE 118-bus system. With 100 correction iterations, screened HardNet++ retains on average 21.5 of 372 flow constraints per iteration and achieves constraint violations and objective quality comparable to an unscreened 500-iteration model, while reducing inference time from 34.4 ms to 6.1 ms per instance. These results demonstrate that adaptive screening can substantially improve the scalability of iterative hard constraint enforcement without sacrificing feasibility or solution quality.
\end{abstract}


\section{Introduction}
Neural networks offer a way to amortize repeated evaluations of computationally expensive simulators or optimization solvers across a broad range of domains by learning a direct map from problem instances to candidate solutions. In many simulation, control, and decision-making applications, however, a useful prediction must satisfy more than an accuracy or objective-value criterion. The output must also satisfy physical laws, safety requirements, or operational limits. Regularization-based approaches include a penalty on constraint violations in the loss function to encourage constraint satisfaction, but do not guarantee feasibility at inference time in general. \emph{Hard-constrained} neural networks address this gap by enforcing constraints by construction.

A common approach is to augment a constraint enforcement layer at the output of the raw neural network that corrects the otherwise-infeasible output to meet the constraints. Certain constraints possess exploitable structure for which enforcement can be implemented with a closed-form projection onto the feasible set~\citep{min2024,chen2024physics,goertzen2025eco}. For problems that lack the special structure admitting a single closed-form projection, iterative methods are often used for constraint enforcement. PiNet~\citep{grontas2026pinet} decomposes a convex feasible set into the intersection of two sets and iteratively projects onto each using the Douglas-Rachford operator splitting approach. DC3~\citep{donti2021dc3} satisfies equality constraints through a variable completion step and inequality constraints via gradient-based corrections. ENFORCE~\citep{lastrucci2025enforce} iteratively projects onto local linearizations of nonlinear constraints. FSNet~\citep{nguyen2026fsnet} formulates the constraint satisfaction problem as an optimization problem to minimize aggregate constraint violation. This paper builds on \name~\citep{goertzen2026hardnetpp}, which uses damped local linearizations to provably drive violations of both nonlinear equality and inequality constraints to zero under suitable regularity conditions, even when the feasible set is nonconvex.

Although iterative constraint-enforcement layers extend hard-constrained learning to more general constraint sets, they can incur substantial computational and memory costs. For a problem with $m$ constraints and $n$ decision variables, each correction step typically involves evaluating constraint residuals and Jacobians and computing an update based on them. Repeating these operations over $K$ iterations can dominate the cost of a neural-network forward pass as the problem size grows. This burden is particularly significant for learned optimization surrogates with many constraints. To offer a practical advantage over conventional solvers, these models must achieve reliable constraint satisfaction without sacrificing their computational efficiency. This challenge is especially relevant in applications such as optimal power flow (OPF), where grid operators must repeatedly determine generator dispatches as loads, renewable generation, and other operating conditions evolve. Moreover, uncertainty in these quantities may require evaluating multiple operating scenarios within a single decision window, making inference latency particularly important. These challenges motivate methods that reduce computational costs by selectively omitting inequality constraints that are sufficiently far from their bounds.

To address these gaps, we propose \emph{Screened \name}. Our work extends \name~\citep{goertzen2026hardnetpp} with a batchwise applied constraint screening method to reduce correction costs in problems with many constraints. In applications such as OPF, only a small subset of inequality constraints is active or near-active at a given operating point. Prior learning-based methods~\citep{hasan2021hybrid,deka2019learning} exploit this observation by predicting active sets to simplify the optimization problem. In this work, we leverage it to reduce the cost of iterative constraint enforcement. At each iteration, our approach retains equality constraints and screens out inequalities only when they are sufficiently far from their bounds across the batch. A screening margin controls how aggressively constraints are omitted. We show that, under the same regularity assumptions as the original \name, the proposed method retains its exponential decay guarantee for constraint violations.

Furthermore, we establish universal approximation at any fixed iteration number under, for arbitrary batch sizes and screening policies. We then extend the iteration-uniform approximation-error transfer guarantees of \citet{bae2026expressivity} to batchwise screening. This bounds the amplification of raw-network approximation error independently of the iteration count and batch size, ensuring that a single raw network with a sufficient parameter budget can achieve a prescribed approximation accuracy after correction, independent of the correction depth. The same guarantee extends to the feasible limiting correction.

Experiments on the IEEE 118-bus AC-OPF problem demonstrate the practical benefits of screening. With 100 correction iterations, our method retains only 21.5 of 372 branch-flow constraint rows per iteration on average and achieves constraint violations and objective quality comparable to the unscreened \name\ with 500 iterations, while reducing inference time from 34.4 ms to 6.1 ms per instance.

\paragraph{Notation and conventions.}
For our approximation results, let $\mathcal X\subset\mathbb R^d$ be a nonempty compact input domain. We denote by $C(\mathcal X;\mathbb R^n)$ and $\mathcal B(\mathcal X;\mathbb R^n)$ the spaces of continuous and bounded functions, respectively, equipped with the uniform norm $\|h\|_\infty:=\sup\limits_{x\in\mathcal X}\|h(x)\|_2$. Vector norms are Euclidean and matrix norms are spectral, both denoted by $\|\cdot\|_2$. For a batch $\mathbf x=(x_1,\ldots,x_B)\in\mathcal X^B$ and a bounded batch-valued map $F=(F_1,\ldots,F_B)$, we use the block-uniform norm $\|F\|_{\infty,\max,B}:=\sup\limits_{\mathbf x\in\mathcal X^B}\max\limits_{1\le i\le B}\|F_i(\mathbf x)\|_2$. All density and closure statements are understood in the corresponding uniform topologies. We write $\mathbb N_0=\{0,1,\ldots\}$, $I_p$ for the $p\times p$ identity matrix, and $\preceq$ for the Loewner order on symmetric matrices. Vector inequalities and $\mathrm{ReLU}$ are understood componentwise. For input-dependent constraints $c_x$, we write $J_c(x,y)=D_yc_x(y)$ for the Jacobian with respect to $y$, whenever it exists, and suppress the input dependence when unambiguous.

\section{\name\ Background}\label{sec:hnpp_background}
In our recent work~\citep{goertzen2026hardnetpp}, we introduced \name, a methodology for nonlinear constraint enforcement in neural networks. Consider a neural network parameterized by $\theta$ that maps an input $x\in\mathbb{R}^d$ to an unconstrained output $\hat{y}_\theta\in\mathbb{R}^n$. 
The goal of \name\ is to move the output toward feasibility through an iterative update scheme. The target feasible set is 
\begin{equation}\label{eq:constraint_set}
    C(x):=
    \left\{y\in\mathbb{R}^n:
    b^l_x\leq c_x(y)\leq b^u_x\right\}.
\end{equation}

Here, $c_x:\mathbb{R}^n\rightarrow\mathbb{R}^m$ represents $m$ scalar constraint functions with upper and lower bounds $b^l_x, b^u_x\in\mathbb{R}^m$. Equality constraints are included by setting $b^l_i=b_i^u$ for the $i$th constraint.

A \name\ iterate is computed based on a local linearization of the constraint, with an additional damping term. Define the constraint Jacobian $J_{c,k}:=J_c(y_k)$ and the constraint residual 
\begin{equation}\label{eq:residual}
r(y):= \operatorname{ReLU}\left(c(y)-b^u\right)-\operatorname{ReLU}(b^l-c(y))
\end{equation}

where $\operatorname{ReLU}$ is applied elementwise. Initialize $y_0=\hat{y}_\theta$. An update step is performed as
\begin{equation}\label{eq:hnpp_step}
    y_{k+1}^{\textsf{HardNet++}} = y_k -J_{c,k}^\top (J_{c,k}J_{c,k}^\top +\varepsilon I)^{-1}r(y_k), \quad \varepsilon>0
\end{equation}

After $K$ iterations, the constrained network output is $\bar{y}_\theta:=y_K^{\textsf{HardNet++}}$. The damping term ensures that the linear system is well-defined even when $J_{c,k}$ is rank-deficient or there are more constraints than variables. Importantly, the use of $J_{c,k}$ distinguishes \name\ from a Levenberg-Marquardt step for minimizing the violation energy $V(y) = \frac12||r(y)||^2$. A Levenberg-Marquardt step instead uses the residual Jacobian $J_{r,k}:=J_r(y_k)$:
\begin{equation}\label{eq:LM_step}
    y_{k+1}^\text{LM} = y_k -J_{r,k}^\top (J_{r,k}J_{r,k}^\top +\varepsilon I)^{-1}r(y_k), \quad \varepsilon>0.
\end{equation}

For an inequality strictly inside its bounds, the corresponding row of $J_r$ is zero, and when a bound is violated, the corresponding row of $J_r$ equals the corresponding row of $J_c$. Thus, Levenberg--Marquardt ignores the local geometry of satisfied inequality constraints. In contrast, \name\ retains every row of $J_c$, which is equivalent to locally treating each satisfied inequality as an equality constraint at its current value. Damping relaxes this local-equality condition, allowing limited motion normal to satisfied constraint boundaries while still biasing the update toward preserving their satisfaction. 
The original analysis further shows that, under appropriate regularity conditions, the full-constraint iteration drives constraint violations arbitrarily close to zero.

\subsection{Computational scaling}
The efficiency of \name\ depends on both the cost of a given iteration~\eqref{eq:hnpp_step} as well as the total number of iterations. Recall that $J_{c,k}\in\mathbb{R}^{m\times n}$ and let $r_k:=r(y_k)$, and $p:=\min\{m,n\}$. The update admits the equivalent forms $J_{c,k}^\top(J_{c,k}J_{c,k}^\top + \varepsilon I_m)^{-1}r_k = (J_{c,k}^\top J_{c,k} + \varepsilon I_n)^{-1}J_{c,k}^\top r_k$ so it can be computed using the smaller of the $m\times m$ and $n\times n$ regularized Gram systems. Forming the resulting $p\times p$ Gram matrix costs $\mathcal{O}(mnp)$, and its Cholesky factorization costs $\mathcal{O}(p^3)$. The matrix-vector products and triangular solves do not change this leading-order cost. Thus, one correction step costs $\mathcal{O}(p^3+mnp)$. For general nonlinear constraints, $J_{c,k}$ changes across iterations, so the Gram matrix and its factorization cannot generally be reused. Therefore, $K$ correction steps cost $\mathcal{O}\!\left(K(p^3+mnp)\right)$.

The total number of iterations can also increase with the number of constraints. Let $V(y):=\frac{1}{2}\|r(y)\|_2^2$, and let $G$, $L$, and $\mu$ denote, respectively, a Lipschitz constant of the constraint map $c$, a smoothness constant of $V$, and the P\L{} constant of $V$. Under the assumptions of~\cite{goertzen2026hardnetpp}, if a feasible point $y^\star$ exists and $\varepsilon>L/2$, then $V(y_k)\leq\bigl(1-2\mu \gamma(\varepsilon)\bigr)^k V(y_0)$. For a target violation-energy tolerance $0<\tau<V(y_0)$, any iteration budget satisfying $K\geq K_\tau^\mathrm{cert}:=\frac{log(\tau/V(y_0))}{log(1-2\mu\gamma(\varepsilon))}$ guarantees $V(y_K)\leq\tau$. Here, $\gamma(\varepsilon)=\min\left\{
\frac{1}{\varepsilon}-\frac{L}{2\varepsilon^2},
\;
\frac{1}{\varepsilon+G^2}-\frac{L}{2(\varepsilon+G^2)^2}
\right\}$. Although $K_\tau^\mathrm{cert}$ is not an explicit function of $m$ alone, it can grow as the number of constraints increases. For a fixed absolute tolerance $\tau$, additional constraints with nonnegligible initial violations increase $V(y_0)$. Moreover, the constants $G$, $L$, and $\mu$ depend on the geometry of the full stacked constraint system and can deteriorate as additional nonlinear constraints are introduced. For example, if additional constraints result in a larger $G$ or $L$, or smaller $\mu$, the certified iteration budget increases. Thus, larger constraint sets can increase both the cost per iteration and the number of iterations required to attain a prescribed feasibility tolerance.

\section{Proposed method: Screened \name}\label{sec:screening}
\cite{goertzen2026hardnetpp} highlight the contrast between two extremes: (1) the Levenberg-Marquardt case where only the violated constraints are considered and (2) the original \name\ approach, where the geometry of every constraint is considered. By including the geometry of all the constraints in the projection step, that is, by using $J_c$ instead of $J_r$ in \eqref{eq:hnpp_step}, a \name\ step limits movement in the direction toward constraint violation. However, there are cases in which only a small subset of inequality constraints are active at an optimum. For example, \citet[Table~2]{hasan2021hybrid} demonstrate empirically for AC-OPF that on average, less than 5\% of the voltage and branch constraints are active across sampled loading conditions for systems ranging from 39 to 1354 buses. When a neural network raw output $\hat y_\theta$ is at or near the constraints for only a small subset of the inequality constraints, while the majority of inequality constraints are comfortably satisfied, using the full $J_c$ in a \name\ step can be wasteful. That is, restricting movement toward constraint violations in a given step may not be necessary for constraints that are far away from being violated. In this case, including those constraints in the construction of $J_c$ increases computational burden without providing meaningful improvements in constraint satisfaction. This motivates a screened approach, balancing the two extremes of considering all constraints at all iterations and only considering violated constraints.

Consider a one-sided inequality $c_j(y)\le b_j^u$. For a screening
parameter $0\le\delta\le1$ and a scale $s_j>0$, retain row $j$ when
$c_j(y_k)-b_j^u\ge-\delta s_j$. For a lower bound, the analogous test is
$b_j^l-c_j(y_k)\ge-\delta s_j$. For a two-sided inequality with
$b_j^l<b_j^u$, take $s_j=(b_j^u-b_j^l)/2$ and retain its row when it is
close to either boundary or violates either bound:
\begin{equation}\label{eq:screening_test}
 \max\{b_j^l-c_j(y_k),\ c_j(y_k)-b_j^u\}\ge-\delta s_j.
\end{equation}
Let $\mathcal S_{\mathrm{always}}$ contain the equality rows and any
inequalities not subject to screening, and let $\mathcal S_k$ be the union
of these rows with the inequality rows retained by the tests above.
The row-selection matrix $P_k$ consists of the corresponding distinct rows
of $I_m$. The screened Jacobian is $J_k^\delta=P_kJ_{c,k}$, and the
residual must be restricted to the same rows.

At $\delta=0$, strictly interior inequalities are discarded, while
violated inequalities, boundary-active inequalities, and all equalities
are retained. Away from inequality boundaries, this selects the same
nonzero rows used by the residual-Jacobian step. At $\delta=1$, every
two-sided row with the half-width scale above is retained. This
full-Jacobian endpoint need not hold for one-sided inequalities with an
arbitrary physical scale. Increasing $\delta$ enlarges the retained set
and incorporates more of the geometry of satisfied constraints.

\subsection{Implementation of Screened \name}
We now introduce \emph{Screened HardNet++}, an adaptive constraint-enforcement procedure that reduces the computational burden of HardNet++ by retaining only selected constraint rows at each correction step. For each sample, the retained inequalities are determined by the screening tests above. We then form a common selection across the batch and apply the HardNet++ correction using the resulting reduced Jacobians. The update is defined below and summarized in Algorithm~\ref{alg:screened_hnpp}.

For a batch $\mathbf x$, write $r_{x_i}$ and
$J_c(x_i,\cdot)$ to restore the input dependence. Let $\mathcal S_{i,k}$
contain the screened inequality rows retained by the local test for sample
$i$ at $y_{i,k}$. We use a common row selecting matrix $P_k$ for the batch, whose rows
are indexed by
\begin{equation}\label{eq:batch_union}
 \mathcal S_k^{\mathrm{batch}}
 =\mathcal S_{\mathrm{always}}\cup\bigcup_{i=1}^B\mathcal S_{i,k}.
\end{equation}
Starting at $y_{i,0}=\hat y_\theta(x_i)$, the screened updates are
\begin{equation}\label{eq:screened_batch_step}
\begin{aligned}
 \tilde J_{i,k}&=P_kJ_c(x_i,y_{i,k}),&
 \tilde r_{i,k}&=P_kr_{x_i}(y_{i,k}),\\
 y_{i,k+1}&=y_{i,k}-\tilde J_{i,k}^\top
 (\tilde J_{i,k}\tilde J_{i,k}^\top+\varepsilon I)^{-1}\tilde r_{i,k},
 &\varepsilon&>0.
\end{aligned}
\end{equation}

\begin{algorithm2e}[t]
\caption{Screened \name}
\label{alg:screened_hnpp}

\KwIn{Batch $(x_i)_{i=1}^{B}$, raw network $\hat y_\theta$,
margin $\delta$, damping $\varepsilon>0$, iteration count $K$}
\KwOut{Corrected predictions $(y_{i,K})_{i=1}^{B}$}

$y_{i,0}\gets \hat y_\theta(x_i)$ for $i=1,\ldots,B$\;

\For{$k=0,\ldots,K-1$}{
  $\mathcal S_{i,k}\gets
  \operatorname{Screen}_\delta(x_i,y_{i,k})$
  for all $i$\;

  $\mathcal S_k^{\mathrm{batch}}\gets
  \mathcal S_{\mathrm{always}}
  \cup\bigcup_{i=1}^{B}\mathcal S_{i,k}$\;

  \If{$\mathcal S_k^{\mathrm{batch}}=\varnothing$}{
    $y_{i,k+1}\gets y_{i,k}$ for all $i$;
    \textbf{continue}\;
  }

  \For{$i=1,\ldots,B$}{
    $\tilde J_{i,k}\gets
    J_c(x_i,y_{i,k})[\mathcal S_k^{\mathrm{batch}},:]$,
    $\tilde r_{i,k}\gets
    r_{x_i}(y_{i,k})[\mathcal S_k^{\mathrm{batch}}]$\;

    Solve
    $(\tilde J_{i,k}\tilde J_{i,k}^{\top}
    +\varepsilon I)z_{i,k}=\tilde r_{i,k}$\;

    $y_{i,k+1}\gets
    y_{i,k}-\tilde J_{i,k}^{\top}z_{i,k}$\;
  }
}
\end{algorithm2e}

An empty selection gives a zero update. Since the local screening tests
retain every violated inequality and all equality rows are always kept,
\begin{equation}\label{eq:screening_residual_preservation}
 P_k^\top P_kr_{x_i}(y_{i,k})=r_{x_i}(y_{i,k})\qquad\text{for every }i.
\end{equation}
For continuously differentiable $c_x$, set
$V_x(y)=\frac12\|r_x(y)\|_2^2$. The full violation gradient is preserved:
\[
 \tilde J_{i,k}^\top\tilde r_{i,k}
 =J_c(x_i,y_{i,k})^\top r_{x_i}(y_{i,k})
 =\nabla V_{x_i}(y_{i,k}).
\]
The preconditioner
$(\tilde J_{i,k}^\top\tilde J_{i,k}+\varepsilon I)^{-1}$ can nevertheless
depend on the other samples, so the corrected output for a given input can
depend on its batch companions.

\subsection{Scalability of Screened \name}
The screened approach reduces the cost of the \name\ iterates. Let
$m_{\delta,k}:=|\mathcal S_k^{\mathrm{batch}}|$ denote the number of
constraints retained at iteration $k$ under union screening, and define
$p_{\delta,k}:=\min\{m_{\delta,k},n\}$. The per-sample cost is then reduced
from $\mathcal{O}\!\left(p^3+mnp\right)$, where $p:=\min\{m,n\}$, to $\mathcal{O}\!\left(p_{\delta,k}^3+m_{\delta,k}np_{\delta,k}\right)$. Screening can also affect the number of correction iterations. Removing rows
that impede progress toward unsatisfied constraints may reduce the required
number of iterations. Conversely, an omitted constraint can become violated and
require later correction. The margin $\delta$ controls this tradeoff.
\section{Convergence of Constraint Violations}\label{sec:convergence}

\begin{proposition}[Violation decay under residual-preserving screening] \label{prop:screened_convergence}
Suppose that, for every $x$, $C(x)\ne\varnothing$, the constraint map
$c_x$ is continuously differentiable and $G$-Lipschitz, and
$V_x(y)=\frac12\|r_x(y)\|_2^2$ is $L$-smooth and satisfies
$\|\nabla V_x(y)\|_2^2\ge2\mu V_x(y)$, with constants
$0<\mu<L$ and $G<\infty$ uniform in $x$. For fixed damping
$\varepsilon>L/2$ and selection matrices in~\eqref{eq:screening_residual_preservation}, set
\begin{equation}\label{eq:screened_convergence_rate}
 \gamma(\varepsilon)=\min\left\{
 \frac1\varepsilon-\frac{L}{2\varepsilon^2},\quad
 \frac1{\varepsilon+G^2}-\frac{L}{2(\varepsilon+G^2)^2}
 \right\},\qquad q=1-2\mu\gamma(\varepsilon)\in(0,1).
\end{equation}
Then every sample in~\eqref{eq:screened_batch_step} satisfies
$V_{x_i}(y_{i,k})\le q^kV_{x_i}(y_{i,0})$ for all $k\ge0$.
\end{proposition}
The proof is in Appendix~\ref{app:screened_convergence}. 
\section{Universal Approximation of Screened \name} \label{sec:ua}
The preceding sections address the computational cost of correction and
the reduction of constraint violations as the iteration count increases.
However, scalability also depends on the raw-network capacity needed
to maintain a prescribed approximation accuracy as more correction
steps are taken. Universal approximation at each fixed iteration count
does not address this issue, since the raw approximant required for a
given accuracy may depend on the iteration count.
\citet[Theorems~5 and~9]{bae2026expressivity} establish
iteration-uniform approximation guarantees under global smoothness,
the Polyak--Lojasiewicz (PL) condition, and, for \name,
global Lipschitzness of the constraint functions.
Since these assumptions do not directly cover AC-OPF over unrestricted
outputs, we first establish fixed-iteration universal approximation
under weaker assumptions, explicitly accounting for batchwise screening.
We then extend the iteration-uniform error-transfer guarantee to the
batchwise setting under the stronger assumptions.
Proofs are provided in Appendix~\ref{app:ua}.

\subsection{Universal approximation at a fixed-iteration count} \label{sec:ua_fixed_k}
Define
\begin{equation}\label{eq:ua_feasible}
 C(x)=\{y\in\mathbb R^n:r_x(y)=0\},\qquad
 \mathcal F=\{f\in C(\mathcal X;\mathbb R^n):f(x)\in C(x)\text{ for all }x\}.
\end{equation}
Assume $\mathcal F\ne\varnothing$ and ordered bounds $b_x^l\le b_x^u$.
For finite $K,B$, let $\mathfrak P_{K,B}$ consist of well-defined policies for~\eqref{eq:screened_batch_step}, allowing sample-specific row-selecting matrices $P_{i,k}$ and damping $\varepsilon_{i,k}\ge\underline\varepsilon>0$ to depend on the entire batch history. The row-selecting matrices consist of distinct rows of $I_m$. The common lower bound $\underline\varepsilon$ on damping applies to all samples, batches, and histories. Formal policy details are given in Appendix~\ref{app:ua_policies}. Starting from $y_{i,0}=h(x_i)$, write
\begin{equation}\label{eq:ua_batch_maps}
\begin{aligned}
 (\mathcal T^\pi_{K,B}h)(\mathbf x)&=(y_{1,K},\ldots,y_{B,K}),&
 (\Delta_Bf)(\mathbf x)&=(f(x_1),\ldots,f(x_B)).
\end{aligned}
\end{equation}
The targets are lifts of single-input feasible functions. For $f\in\mathcal F$, assume the following.
\begin{equation}\label{eq:ua_regularity}
 \omega_f(t):=\sup\limits_{\substack{x\in\mathcal X\\\|y-f(x)\|_2\le t}}
 \|r_x(y)\|_2<\infty\quad(0\le t<\infty),
 \qquad\lim_{t\downarrow0}\omega_f(t)=0.
\end{equation}
Joint continuity of the residual map is a sufficient condition for~\eqref{eq:ua_regularity}. Global Lipschitzness of constraint functions~\citet[Theorem~9]{bae2026expressivity} also suffices, giving $\omega_f(t)\le Gt$ without requiring continuity in the input.

\begin{theorem}\label{thm:ua_batch}
Let $f\in\mathcal F$ satisfy~\eqref{eq:ua_regularity} and fix
$K\in\mathbb N_0$. There is a finite nondecreasing modulus $\Phi_{f,K}$,
with $\Phi_{f,K}(0)=0$ and $\Phi_{f,K}(t)\to0$ as $t\downarrow0$, such that
every $h\in C(\mathcal X;\mathbb R^n)$ satisfies
\begin{equation}\label{eq:ua_batch_bound}
 \sup\limits_{B\ge1}\sup\limits_{\pi\in\mathfrak P_{K,B}}
 \|\mathcal T^\pi_{K,B}h-\Delta_Bf\|_{\infty,\max,B}
 \le\Phi_{f,K}(\|h-f\|_\infty).
\end{equation}
Consequently, if the set of raw approximators $\mathcal N\subset C(\mathcal X;\mathbb R^n)$ is dense, then for every $\eta>0$ there exists $h\in\mathcal N$ such that $\sup\limits_{B\ge1}\sup\limits_{\pi\in\mathfrak P_{K,B}} \|\mathcal T^\pi_{K,B}h-\Delta_Bf\|_{\infty,\max,B} < \eta$. Thus for all batch sizes, compositions, and policies, the corrected class is dense in $C(\mathcal X;\mathbb R^n)$, and every target $f\in\mathcal F$ can be approximated arbitrarily closely.
\end{theorem}
The case $B=1$ corresponds to the original formulation of HardNet++~\citep{goertzen2026hardnetpp} without screening. Corrected functions are bounded but may be discontinuous. Appendix~\ref{app:ua_modulus} gives the explicit modulus. Moreover, Proposition~\ref{prop:ua_local} establishes an explicit raw approximation tolerance under local residual growth, which applies to continuous feasible AC-OPF targets without requiring global smoothness or the PL condition; see Lemma~\ref{lem:ua_local_sufficient}.

\subsection{Iteration-uniform error transfer} \label{sec:ua_uniform}
We now strengthen the fixed-iteration result by controlling approximation error uniformly over all finite iteration numbers. Under the assumptions of Proposition~\ref{prop:screened_convergence}, the raw-network approximation error transfers to the corrected model with a factor independent of the iteration count and batch size. This follows by extending \citet[Theorem~5]{bae2026expressivity} to batch-dependent screening.

\begin{corollary} \label{cor:ua_uniform}
Under the assumptions of Proposition~\ref{prop:screened_convergence}, take
$\underline\varepsilon=\varepsilon$. Let
$\mathfrak P^{\mathrm{res}}_{K,B}$ denote policies in $\mathfrak P_{K,B}$
using $J_c=D_yc_x$, the fixed damping, and row selecting matrices retaining every
nonzero residual coordinate. This includes the screened update in~\eqref{eq:screened_batch_step}.
Set $A_\varepsilon=(\varepsilon\gamma(\varepsilon))^{-1}\sqrt{L/\mu}$,
with $\gamma(\varepsilon)$ from~\eqref{eq:screened_convergence_rate}.
For every $f\in\mathcal F$ and nonempty
$\mathcal{N}\subset C(\mathcal X;\mathbb R^n)$, writing
$e_{\mathcal{N}}(f)=\inf\limits_{h\in\mathcal{N}}\|h-f\|_\infty$, we have
\begin{equation}\label{eq:ua_uniform_transfer}
 \inf\limits_{h\in\mathcal{N}}\sup\limits_{K\in\mathbb N_0}\sup\limits_{B\ge1}
 \sup\limits_{\pi\in\mathfrak P^{\mathrm{res}}_{K,B}}
 \|\mathcal T^\pi_{K,B}h-\Delta_Bf\|_{\infty,\max,B}
 \le(1+A_\varepsilon)e_{\mathcal{N}}(f).
\end{equation}
\end{corollary}
Taking $\mathcal{N}=\{h\}$ gives the bound for each raw function. If $\mathcal{N}$ is dense, one raw network achieves any prescribed accuracy simultaneously for every finite $K$, batch, and admissible policy. More generally, raw approximation-error bounds transfer with a factor independent of $K$ and $B$, so a sufficient raw-network parameter budget for a prescribed accuracy can be chosen independently of the iteration count. The factor may still depend on the constraint geometry.


\subsection{Approximation after convergence}
The approximation result in Section~\ref{sec:ua_fixed_k} does not require convergence. Under the  smoothness and PL assumptions used for \name\ in \citet{goertzen2026hardnetpp}, the usual exponential violation-decay bound also extends to the screened update. Under the assumptions of Corollary~\ref{cor:ua_uniform}, fix a batch size $B$ and a single screening policy $\pi$ whose restriction to each finite horizon $K$ belongs to $\mathfrak P^{\mathrm{res}}_{K,B}$. The exponential violation decay in Proposition~\ref{prop:screened_convergence}, together with the step bound \eqref{eq:ua_step_bound}, makes the successive displacements summable. Hence each sample converges to a feasible limit $y_{i,\infty}\in C(x_i)$. Writing
$(\mathcal T^\pi_{\infty,B}h)(\mathbf x)
=(y_{1,\infty},\ldots,y_{B,\infty})$,
taking $\mathcal{N}=\{h\}$ in Corollary~\ref{cor:ua_uniform} and passing
to the limit gives, for every $f\in\mathcal F$,
\[
 \|\mathcal T^\pi_{\infty,B}h-\Delta_B f\|_{\infty,\max,B}
 \le (1+A_\varepsilon)\|h-f\|_\infty.
\]
Consequently, if the raw network class is uniformly dense, the converged
outputs approximate any continuous feasible target to arbitrary accuracy,
using a single raw network uniformly over batch sizes, compositions,
and such policies.

\section{Experiments}
\subsection{AC-OPF}
We evaluate our screened \name\ approach on the AC optimal power flow (OPF) problem. See Appendix~\ref{sec:acopf} for a full description of the problem. We use the \texttt{pglib\_opf\_case118\_ieee} instance from PGLib-OPF~\citep{babaeinejadsarookolaee2019power}, with 118 buses, 54 of which are generators, and 186 branches. Each input instance specifies the  active and reactive power demand at all buses. The decision variables are the active and reactive power generation at each generator and the voltage magnitudes $v$ and angles $\theta$ at each bus. The goal is to learn a mapping from the demand instance ($x\in\mathbb{R}^{236}$) to the optimal decision variables ($y\in\mathbb{R}^{344}$) that satisfy the constraints in~\eqref{eq:active_balance}-\eqref{eq:thermal_limits}. In total, there are 835 constraints, so an unscreened \name\ step forms an $835\times344$ Jacobian. These constraints are summarized in Table~\ref{tab:case118-constraints} in Appendix~\ref{app:opf_details}. We apply screening only to the branch-flow inequality constraints, which bound the apparent-power flow at each end of a transmission branch.

We take an unsupervised training approach in which we use the objective function~\eqref{eq:acopf_obj} directly as the loss function for training. The network is a MLP with two hidden layers of width 256. Evaluation is performed on 200 unseen instances and compared against an IPOPT~\citep{wachter2006implementation} solution. Inference was performed on a single NVIDIA Tesla V100 GPU (32\,GB), with all 200 test instances in a single batch. IPOPT solved the same instances individually on 2 cores of the CPU on the same node.

\begin{table}[th]
  \centering
  \caption{AC-OPF on \texttt{pglib\_opf\_case118\_ieee}. Flow rows kept refers to the average number of the 372 branch-flow rows used per projection iteration. $J_c$ and $J_r$ represent the choice of Jacobian for the update step. The max constraint violation is across all constraints and instances.}
  \label{tab:screening}
  \footnotesize
  \setlength{\tabcolsep}{3pt}
  \renewcommand{\arraystretch}{0.9}
  \begin{tabular}{@{}lrrrr@{}}
    \toprule
    & \multicolumn{2}{c}{500 iterations} & \multicolumn{2}{c}{100 iterations} \\
    \cmidrule(lr){2-3} \cmidrule(lr){4-5}
    & $J_r$ & $J_c$ & $J_c$ & $J_c$ \\
    \midrule
    Flow screening & -- & -- & $\delta = 0.2$ & none \\
    Flow rows kept & 372 & 372 & 21.5 & 372 \\
    \addlinespace
    \textit{Suboptimality (\%)} \\
    \quad mean & $2.11$ & $0.20$ & $0.18$ & $-2.94$ \\
    \quad max & $4.21$ & $0.44$ & $0.35$ & $-2.06$ \\
    Cheaper than IPOPT & 0/200 & 0/200 & 0/200 & 200/200 \\
    \addlinespace
    \textit{Equality violation} \\
    \quad max & $3.9\times10^{-4}$ & $2.2\times10^{-3}$ & $2.3\times10^{-3}$ & $9.8\times10^{-2}$ \\
    \quad mean & $2.3\times10^{-6}$ & $2.1\times10^{-4}$ & $2.3\times10^{-4}$ & $1.8\times10^{-2}$ \\
    \addlinespace
    \textit{Inequality violation} \\
    \quad max & $5.8\times10^{-3}$ & $2.6\times10^{-2}$ & $2.2\times10^{-2}$ & $8.8\times10^{-2}$ \\
    \quad mean & $9.2\times10^{-7}$ & $1.4\times10^{-4}$ & $1.1\times10^{-4}$ & $3.4\times10^{-3}$ \\
    \addlinespace
    Time (ms) & 33.4 & 34.4 & 6.1 & 7.5 \\
    \bottomrule
  \end{tabular}
\end{table}

Table~\ref{tab:screening} shows that screening can recover the feasibility and objective quality of full-constraint projections with a substantially smaller iteration budget. With $\delta=0.2$, the 100-iteration screened-$J_c$ model retains only 21.5 of the 372 flow rows per iteration on average, yet achieves suboptimality and constraint violations comparable to the 500-iteration full-$J_c$ model. This reduces inference time from 34.4\,ms to 6.1\,ms per instance. In contrast, the 100-iteration full-$J_c$ model remains appreciably infeasible, as reflected in its negative reported suboptimality, while the 500-iteration $J_r$ model attains low violations but a larger objective gap. Thus, screening preserves the useful local geometry of near-active constraints while allowing feasibility to be reached with fewer correction layers. All approaches provide significant improvements in efficiency relative to the IPOPT solver, which took $11.4$ s (11400 ms) on average.

The 500-iteration $J_r$ model attains low violations but has higher suboptimality. Near an OPF optimum, inexpensive generators often operate at or near their limits. When a generator-limit inequality is satisfied, its residual and corresponding $J_r$ row are zero, so updates that resolve other violations do not account for preserving that near-active limit. They can consequently shift generation away from inexpensive units and toward more costly dispatches. By retaining the geometry of near-active constraints, the screened-$J_c$ approach avoids this loss in objective quality while requiring far fewer iterations than full-$J_c$.

\begin{figure}[t]
    \centering
        \includegraphics[width=0.8\linewidth]{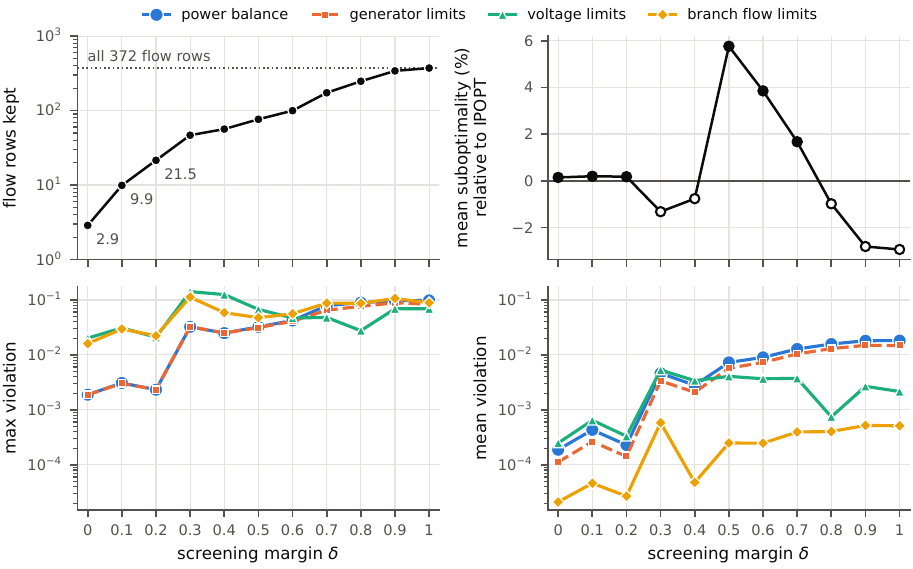}
        \label{fig:panel_one}
    \caption{Retained flow constraints, mean and maximum constraint violations, and suboptimality relative to IPOPT solution for screened \name\ AC-OPF surrogate for choices of screening margin $\delta$. All models use $K=100$ iterates and $\varepsilon=0.1$}
    \label{fig:opf_fig}
\end{figure}

Figure~\ref{fig:opf_fig} highlights a useful screening regime: for $\delta \leq 0.2$, the model retains only a small subset of flow rows while achieving low violations and near-zero suboptimality. At larger margins, performance generally worsens, with violations increasing even for constraint classes that are always retained. This behavior reflects the coupling of the AC-OPF constraints: the selected flow rows influence the full projection direction. The trends are not strictly monotonic because each margin is trained separately, but the result at $\delta=1$ is consistent with Table~\ref{tab:screening}: incorporating all flow-constraint geometry requires more than 100 iterations to recover feasibility.

\section{Conclusion}
In this paper, we introduced \emph{Screened \name} for scalable constraint enforcement in neural networks. By adaptively screening inequality constraints that are sufficiently inactive, the method avoids unnecessary computations in processing constraints that are unlikely to affect the local trajectory. Our results show that this approach can substantially reduce inference time on AC-OPF while maintaining low violation and near-optimal solutions. The screening margin exposes a useful tradeoff: incorporating more constraint geometry can require more correction iterations, whereas a moderate margin can achieve comparable performance with a much smaller per-iteration system. Future work could adapt this margin automatically and assess the approach on larger constrained learning problems.
\subsection*{Generative AI Use}

Generative AI tools were used for drafting parts of the manuscript and implementing parts of the experiments. The authors reviewed and verified all AI-assisted content and take full responsibility for the final content of the paper.



\acks{The authors thank Baosen Zhang for helpful discussions that informed the formulation of this approach. This work was supported in part by the National Science Foundation under CAREER Award No.~2544396 and Graduate Research Fellowship No.~2141064 and by the MIT-Amazon Science Hub. The authors also acknowledge the MIT SuperCloud and Lincoln Laboratory Supercomputing Center for providing computing resources that have contributed to the results reported within this paper. }

\bibliography{l4dc2027-sample}

@article{babaeinejadsarookolaee2019power,
  title={{The Power Grid Library for Benchmarking AC Optimal Power Flow Algorithms}},
  author={Babaeinejadsarookolaee, Sogol and Birchfield, Adam and Christie, Richard D and Coffrin, Carleton and DeMarco, Christopher and Diao, Ruisheng and Ferris, Michael and Fliscounakis, Stephane and Greene, Scott and Huang, Renke and others},
  journal={arXiv preprint arXiv:1908.02788},
  year={2019}
}

@inproceedings{deka2019learning,
  title={{Learning for DC-OPF: Classifying active sets using neural nets}},
  author={Deka, Deepjyoti and Misra, Sidhant},
  booktitle={2019 IEEE Milan PowerTech},
  pages={1--6},
  year={2019},
  organization={IEEE}
}

@article{wachter2006implementation,
  title={{On the implementation of an interior-point filter line-search algorithm for large-scale nonlinear programming}},
  author={W{\"a}chter, Andreas and Biegler, Lorenz T},
  journal={Mathematical Programming},
  volume={106},
  number={1},
  pages={25--57},
  year={2006},
  publisher={Springer}
}

@article{lastrucci2025enforce,
  title={{ENFORCE: Nonlinear Constrained Learning with Adaptive-depth Neural Projection}},
  author={Lastrucci, Giacomo and Schweidtmann, Artur M},
  journal={arXiv preprint arXiv:2502.06774v3},
  year={2025}
}

@article{hasan2021hybrid,
  title={{Hybrid Learning Aided Inactive Constraints Filtering Algorithm to Enhance AC OPF Solution Time}},
  author={Hasan, Fouad and Kargarian, Amin and Mohammadi, Javad},
  journal={IEEE Transactions on Industry Applications},
  volume={57},
  number={2},
  pages={1325--1334},
  year={2021},
  publisher={IEEE}
}

@article{bae2026expressivity,
  author = {Bae, Juho and Azizan, Navid},
  title = {{When Does Constraint Enforcement Preserve Expressivity?}},
  journal = {arXiv preprint},
  year = {2026}
}

@techreport{zimmerman2019opfderivatives,
  author = {Zimmerman, Ray D.},
  title = {{AC Power Flows, Generalized OPF Costs and their Derivatives using Complex Matrix Notation}},
  institution = {Power Systems Engineering Research Center},
  type = {{MATPOWER} Technical Note},
  number = {2},
  year = {2019},
  note = {Revision 7, June 20, 2019},
  url = {https://matpower.org/docs/TN2-OPF-Derivatives.pdf}
}

@article{min2024,
  title={{HardNet: Hard-Constrained Neural Networks with Universal Approximation Guarantees}},
  author={Min, Youngjae and Azizan, Navid},
  journal={arXiv preprint arXiv:2410.10807},
  year={2024}
}

@article{chen2024physics,
  title={{Physics-Informed Neural Networks with Hard Linear Equality Constraints}},
  author={Chen, Hao and Flores, Gonzalo E Constante and Li, Can},
  journal={Computers \& Chemical Engineering},
  volume={189},
  pages={108764},
  year={2024},
  publisher={Elsevier}
}

@article{goertzen2025eco,
  title={{ECO: Energy-Constrained Operator Learning for Chaotic Dynamics with Boundedness Guarantees}},
  author={Goertzen, Andrea and Tang, Sunbochen and Azizan, Navid},
  journal={arXiv preprint arXiv:2512.01984},
  year={2025}
}

@article{goertzen2026hardnetpp,
  title={{HardNet++: Nonlinear Constraint Enforcement in Neural Networks}},
  author={Goertzen, Andrea and Alim, Kaveh and Min, Youngjae and Azizan, Navid},
  journal={IEEE Control Systems Letters},
  year={2026},
  publisher={IEEE}
}

@inproceedings{donti2021dc3,
  title        = {{DC3: A Learning Method for Optimization with Hard Constraints}},
  author       = {Donti, Priya L. and Rolnick, David and Kolter, J. Zico},
  booktitle    = {International Conference on Learning Representations},
  year         = {2021},
  eprint       = {2104.12225},
  archivePrefix= {arXiv},
  primaryClass = {cs.LG}
}

@article{nguyen2026fsnet,
  title={{FSNet: Feasibility-Seeking Neural Network for Constrained Optimization with Guarantees}},
  author={Nguyen, Hoang and Donti, Priya},
  journal={Advances in Neural Information Processing Systems},
  volume={38},
  pages={39670--39708},
  year={2026}
}

@inproceedings{grontas2026pinet,
  title={{Pinet: Optimizing hard-constrained neural networks with orthogonal projection layers}},
  author={Grontas, Panagiotis D and Terpin, Antonio and Balta, Efe and D'Andrea, Raffaello and Lygeros, John},
  booktitle={International Conference on Learning Representations},
  volume={2026},
  pages={62933--62979},
  year={2026}
}

\appendix
\section{Proofs and Supplementary Results for Universal Approximation}
\label{app:ua}

\subsection{Policy conventions and target spaces}
\label{app:ua_policies}
In $\mathfrak P_{K,B}$, selections and damping may depend on the input batch, all current states, and the preceding history. Any auxiliary parameters and random seeds are fixed as part of the policy, so that it defines a deterministic map for each batch. Its updates must be defined on every batch history and obey the common lower bound on the damping $\underline\varepsilon>0$. An empty selection gives a zero update. For ordered bounds, the constraint set $C(x)$ in~\eqref{eq:constraint_set} is exactly the zero set of $r_x$ and one-sided constraints use only the corresponding residual term. A raw network is $h=\hat y_\theta$. The target class consists of diagonal lifts $\Delta_Bf$, not arbitrary batch-dependent feasible maps. Corrected maps may be discontinuous and are compared with continuous feasible targets in the uniform norm on bounded functions.

\subsection{A uniform bound for a damped correction step}
\begin{lemma}[Damped inverse bound]\label{lem:ua_damped_inverse}
For every real rectangular matrix $A$ and $\varepsilon>0$,
\begin{equation}\label{eq:ua_damped_inverse}
 \|A^\top(AA^\top+\varepsilon I)^{-1}\|_2
 \le\frac{1}{2\sqrt\varepsilon}.
\end{equation}
Consequently, a screened step with $A=PJ_c(x,y)$ and
$\varepsilon\ge\underline\varepsilon>0$ satisfies
\begin{equation}\label{eq:ua_step_bound}
 \|y^+-y\|_2\le\beta\|r_x(y)\|_2,
 \qquad\beta=\frac{1}{2\sqrt{\underline\varepsilon}}.
\end{equation}
The same statement holds for any real matrix of the appropriate size used in place of $J_c(x,y)$.
\end{lemma}
\begin{proof}
The singular values of $A^\top(AA^\top+\varepsilon I)^{-1}$ are
$\sigma/(\sigma^2+\varepsilon)$, including zeros, where $\sigma$ represents the singular values of $A$. Since
$\sigma^2+\varepsilon\ge2\sqrt\varepsilon\,\sigma$,
\eqref{eq:ua_damped_inverse} follows. A row selecting matrix formed from distinct
rows of the identity satisfies $\|Pv\|_2\le\|v\|_2$. Applying this to the
screened residual, and using $\varepsilon\ge\underline\varepsilon$, proves
\eqref{eq:ua_step_bound}. An empty selection has zero displacement and
satisfies the same inequality.
\end{proof}

\subsection{Residual regularity and finite-iteration moduli}
\label{app:ua_modulus}
For $f\in\mathcal F$ and $t\ge0$, define the target tube
\[
 Q_f(t)=\{(x,y):x\in\mathcal X,\ \|y-f(x)\|_2\le t\}.
\]
Under~\eqref{eq:ua_regularity}, with $\beta$ as
in~\eqref{eq:ua_step_bound}, define
\begin{equation}\label{eq:ua_phi}
 \Phi_{f,0}(t)=t,\qquad
 \Phi_{f,k+1}(t)=\Phi_{f,k}(t)+\beta\omega_f(\Phi_{f,k}(t)).
\end{equation}
Here $\omega_f(t)$ bounds the residual within distance $t$ of the target,
and $\Phi_{f,k}(t)$ bounds the target error after $k$ corrections starting
within distance $t$.

\begin{lemma}[Residual regularity]
\label{lem:ua_regular}
If $(x,y)\mapsto r_x(y)$ is jointly continuous on
$\mathcal X\times\mathbb R^n$, then~\eqref{eq:ua_regularity} holds for
every $f\in\mathcal F$. Under~\eqref{eq:ua_regularity}, each
$\Phi_{f,k}$ in~\eqref{eq:ua_phi} is finite and nondecreasing on
$[0,\infty)$, satisfies $\Phi_{f,k}(0)=0$, and obeys
$\lim\limits_{t\downarrow0}\Phi_{f,k}(t)=0$ for every fixed finite $k$.
\end{lemma}
\begin{proof}
For finite $t\ge0$, the tube $Q_f(t)$ is compact, since
$\mathcal X$ is compact and $f$ is continuous. Joint
continuity of the residual makes $\omega_f(t)$ finite. Uniform continuity
on $Q_f(1)$, together with $r_x(f(x))=0$, gives
$\omega_f(t)\to0$ as $t\downarrow0$.
More generally, under~\eqref{eq:ua_regularity}, the definition of
$\omega_f$ shows that it is nondecreasing and $\omega_f(0)=0$.
The recursion~\eqref{eq:ua_phi} then proves finiteness, monotonicity,
and the value at zero by induction. The same induction, using the
right-hand limit of $\omega_f$ at zero, gives
$\Phi_{f,k}(t)\to0$ for each fixed finite $k$.
\end{proof}

\paragraph{Relation to a global Lipschitz bound of constraint functions.}
For fixed $x$, the signed residual map
$R_x(z)=\mathrm{ReLU}(z-b_x^u)-\mathrm{ReLU}(b_x^l-z)$ is
nonexpansive. This follows coordinatewise, since each scalar map is
continuous and piecewise affine with slopes in $\{0,1\}$.
The corresponding one-sided and equality residuals have the same property.
Since $R_x(c_x(f(x)))=0$,
\begin{equation}\label{eq:ua_residual_comparison}
 \|r_x(y)\|_2
 \le\|c_x(y)-c_x(f(x))\|_2.
\end{equation}
Thus a uniform bound $\|c_x(u)-c_x(v)\|_2\le G\|u-v\|_2$ gives
$\omega_f(t)\le Gt$, with no continuity assumption on the input dependence.

\subsection{Proof of Theorem~\ref{thm:ua_batch}}
Fix a batch, an admissible policy, and a raw function $h$. For sample $i$,
put $e_{i,k}=\|y_{i,k}-f(x_i)\|_2$. By
Lemma~\ref{lem:ua_damped_inverse} and the definition of $\omega_f$,
\begin{equation}\label{eq:ua_error_recurrence}
 e_{i,k+1}
 \le e_{i,k}+\beta\|r_{x_i}(y_{i,k})\|_2
 \le e_{i,k}+\beta\omega_f(e_{i,k}).
\end{equation}
The right-hand side depends on this sample's approximation error, regardless of
how its selected matrix depends on the other samples or on the history.
Monotonicity in~\eqref{eq:ua_phi} yields, by induction,
\begin{equation}\label{eq:ua_samplewise}
 e_{i,k}\le\Phi_{f,k}(e_{i,0})
 \le\Phi_{f,k}(\|h-f\|_\infty),\qquad 0\le k\le K.
\end{equation}
Taking the maximum over samples and the suprema over batch inputs,
batch sizes, and policies proves~\eqref{eq:ua_batch_bound}.
The finite right-hand side and boundedness of $f$ also give
\[
 \sup\limits_{B\ge1}\sup\limits_{\pi\in\mathfrak P_{K,B}}
 \|\mathcal T^\pi_{K,B}h\|_{\infty,\max,B}
 \le\|f\|_\infty+\Phi_{f,K}(\|h-f\|_\infty)<\infty.
\]
For any $\eta>0$, Lemma~\ref{lem:ua_regular} implies existence of a $t_0>0$ with $\Phi_{f,K}(t_0)<\eta$. Density of $\mathcal{N}$ implies existence of an $h\in\mathcal{N}$ with $\|h-f\|_\infty<t_0$. This choice is made before choosing the batch or
policy, so the same $h$ satisfies
\begin{equation}\label{eq:ua_batch_density}
 \sup\limits_{B\ge1}\sup\limits_{\pi\in\mathfrak P_{K,B}}
 \|\mathcal T^\pi_{K,B}h-\Delta_Bf\|_{\infty,\max,B}<\eta.
\end{equation}
\hfill$\blacksquare$

\paragraph{\name\ as a special case.}
For $B=1$, write $T_K^\pi=\mathcal T^\pi_{K,1}$. A policy can depend on $x$ and the entire local history
$(y_0,\ldots,y_k)$. Equation~\eqref{eq:ua_samplewise} gives
$\sup\limits_{\pi\in\mathfrak P_{K,1}}\|T_K^\pi h-f\|_\infty
\le\Phi_{f,K}(\|h-f\|_\infty)$.
Consequently, for every fixed such policy,
$f$ lies in the uniform closure of $T_K^\pi(\mathcal{N})$ within
$\mathcal B(\mathcal X;\mathbb R^n)$.
For general $B$, the concatenated Euclidean norm obeys
$\|(v_1,\ldots,v_B)\|_2\le\sqrt B\max_i\|v_i\|_2$,
so the batch-size-independent estimate uses the block-maximum norm.

\subsection{Proof of Corollary~\ref{cor:ua_uniform}}
\label{app:ua_uniform}
Fix a finite $K\ge1$, a batch, a policy, an initialization, and a sample
$i$. Write $J_{i,k}=J_c(x_i,y_{i,k})$ and
$M_{i,k}=P_{i,k}^\top P_{i,k}$ so that $M_{i,k}r_{x_i}(y_{i,k})=r_{x_i}(y_{i,k})$. Since $\nabla V_x(y)=J_c(x,y)^\top r_x(y)$, the screened update becomes
\begin{equation}\label{eq:ua_uniform_preconditioned}
 y_{i,k+1}=y_{i,k}-B_{i,k}\nabla V_{x_i}(y_{i,k}),\qquad
 B_{i,k}=(J_{i,k}^\top M_{i,k}J_{i,k}+\varepsilon I)^{-1}.
\end{equation}
The global Lipschitz bound for the differentiable constraint map gives
$\|J_{i,k}\|_2\le G$. Together with $0\preceq M_{i,k}\preceq I$, this yields
\begin{equation}\label{eq:ua_uniform_spectrum}
 aI_n\preceq B_{i,k}\preceq bI_n,\qquad
 a=\frac1{\varepsilon+G^2},\quad b=\frac1\varepsilon<\frac2L.
\end{equation}
These bounds do not depend on the batch, policy, initialization, or step.

To apply \citet[Theorem~5]{bae2026expressivity} to a history-dependent
policy, freeze the realized matrices $B_{i,0},\ldots,B_{i,K-1}$ and treat
them as prescribed preconditioners independent of the state and input.
Extend this sequence by $aI_n$ for subsequent indices. The auxiliary
iteration at input $x_i$, started at $h(x_i)$, agrees with the actual
sample trajectory through step $K$ by induction. It satisfies the
theorem's spectral hypotheses, with
\[
 c=\min\{a-La^2/2,\ b-Lb^2/2\}=\gamma(\varepsilon),\qquad
 A=\frac bc\sqrt{\frac L\mu}=A_\varepsilon.
\]
The finite-iteration approximationerror bound of that theorem therefore gives
\begin{equation}\label{eq:ua_uniform_samplewise}
 \|y_{i,K}-f(x_i)\|_2
 \le(1+A_\varepsilon)\|h(x_i)-f(x_i)\|_2.
\end{equation}
Although the frozen sequence can vary with the batch and initialization,
the constant depends only on the common spectral and regularity bounds.
The case $K=0$ is immediate.

Taking the maximum over samples and the suprema over batch inputs,
policies, batch sizes, and finite iteration counts gives
\begin{equation}\label{eq:ua_uniform_predictor}
 \sup\limits_{K\in\mathbb N_0}\sup\limits_{B\ge1}
 \sup\limits_{\pi\in\mathfrak P^{\mathrm{res}}_{K,B}}
 \|\mathcal T^\pi_{K,B}h-\Delta_Bf\|_{\infty,\max,B}
 \le(1+A_\varepsilon)\|h-f\|_\infty.
\end{equation}
Taking the infimum over $h\in\mathcal{N}$ proves
\eqref{eq:ua_uniform_transfer}. For a dense class and $\eta>0$, choose
one $h$ with $\|h-f\|_\infty<\eta/(1+A_\varepsilon)$;
\eqref{eq:ua_uniform_predictor} then holds with error below $\eta$ for all
finite iteration counts simultaneously. The finite-horizon bound does
not require consistency of the policies across different iteration counts.\hfill$\blacksquare$

\subsection{Dependence on companion samples}
\begin{corollary}[Sensitivity to batch composition near a feasible target]
\label{cor:ua_companions}
Under the assumptions of Theorem~\ref{thm:ua_batch}, fix an input $x$,
its initial state $u$, and the finite iteration count $K$. Let
$Y_K(x,u;\xi)$ be its terminal state in a batch configuration $\xi$,
which specifies its position in an ordered batch, the companion inputs
and their initial states, and the policy. These choices may be arbitrary,
provided the realized updates use matrices with finite entries and share a common lower bound on the damping. Then
\begin{align}
 \|Y_K(x,u;\xi)-f(x)\|_2
 &\le\Phi_{f,K}(\|u-f(x)\|_2),\label{eq:ua_companion_target}\\
 \|Y_K(x,u;\xi)-Y_K(x,u;\xi')\|_2
 &\le2\Phi_{f,K}(\|u-f(x)\|_2)\label{eq:ua_companion_difference}
\end{align}
for any two such configurations. In particular, $u=f(x)$ remains unchanged
in every admissible configuration.
\end{corollary}
\begin{proof}
The induction~\eqref{eq:ua_error_recurrence}--\eqref{eq:ua_samplewise}
applies to the designated sample alone and uses no bound on the companion
states' approximation errors. This proves~\eqref{eq:ua_companion_target}.
The triangle inequality around $f(x)$ proves
\eqref{eq:ua_companion_difference}. If $u=f(x)$, its residual and hence its
update are zero; induction shows that it remains at $f(x)$.
\end{proof}

\subsection{Permutation equivariance of union screening}
\label{app:ua_permutation}
Suppose the local screening tests use each sample's own data and states,
with the same rule at every batch position, and use the common union
\eqref{eq:batch_union}. With fixed damping, or a damping rule that permutes
with the samples, reordering a batch reorders its corrected outputs.
Indeed, this is true at initialization. If it holds through iteration $k$,
the local retained sets are permuted, their union is unchanged, and the
same selected rows are used for the corresponding samples. Their residuals,
constraint matrices, and damping values are likewise matched, so the
updates preserve the permutation. Induction proves the assertion.
The row-ordering convention may be fixed; permuting selected rows also
leaves the exact update invariant. Arbitrary policies in
$\mathfrak P_{K,B}$ need not be permutation equivariant.

\subsection{Local residual growth and explicit tolerances}
\label{sec:ua_local}\label{app:ua_local}
The next result only needs information within distance $\rho$ of the target.
For a batch, consider histories for which every state of sample $i$ lies in
$\{y:\|y-f(x_i)\|_2\le\rho\}$. Let $\mathfrak P^\rho_{K,B}(f)$ consist of
policies whose selections and finite correction matrices are defined for
every such history and whose damping has the same lower bound $\underline\varepsilon$ there. The result ensures that sufficiently accurate initializations never leave these tubes during the prescribed $K$ steps, and their behavior outside these tubes need not be specified.

\begin{proposition}[Local residual growth and an explicit raw tolerance]
\label{prop:ua_local}
Fix $f\in\mathcal F$ and $K\in\mathbb N_0$. Suppose that for some
$\rho>0$ and $0\le L_f<\infty$,
\begin{equation}\label{eq:ua_local_growth}
 \|r_x(y)\|_2\le L_f\|y-f(x)\|_2
 \quad\text{whenever }x\in\mathcal X,\ \|y-f(x)\|_2\le\rho.
\end{equation}
Put $a_f=1+L_f/(2\sqrt{\underline\varepsilon})$.
If $h\in C(\mathcal X;\mathbb R^n)$ satisfies
$\|h-f\|_\infty\le\rho a_f^{-K}$, then every sample remains in its
target tube through iteration $K$, and
\begin{equation}\label{eq:ua_local_bound}
 \sup\limits_{B\ge1}\ \sup\limits_{\pi\in\mathfrak P^\rho_{K,B}(f)}
 \|\mathcal T^\pi_{K,B}h-\Delta_Bf\|_{\infty,\max,B}
 \le a_f^K\|h-f\|_\infty.
\end{equation}
In particular, a sufficient raw approximation tolerance for corrected
error less than $\eta>0$ is
\begin{equation}\label{eq:ua_local_tolerance}
 \|h-f\|_\infty<a_f^{-K}\min\{\rho,\eta\}.
\end{equation}
\end{proposition}

\paragraph{Proof of Proposition~\ref{prop:ua_local}.}
Fix a batch and a policy in $\mathfrak P^\rho_{K,B}(f)$, and put $e_{i,k}=\|y_{i,k}-f(x_i)\|_2$. Initially, all
samples lie in their respective tubes because
$e_{i,0}\le\|h-f\|_\infty\le\rho a_f^{-K}\le\rho$.
If the batch history stays in the tubes through time $k<K$, the policy
provides finite matrices and the required lower bound on the damping for the next
updates. Lemma~\ref{lem:ua_damped_inverse}
and~\eqref{eq:ua_local_growth} give, for every sample,
\[
 e_{i,k+1}\le e_{i,k}+\beta L_f e_{i,k}=a_f e_{i,k}.
\]
Simultaneous induction over the batch therefore yields
\begin{equation}\label{eq:ua_tube_induction}
 e_{i,k}\le a_f^k e_{i,0}
 \le\rho a_f^{k-K}\le\rho,\qquad 0\le k\le K.
\end{equation}
This proves both that every required update is defined and that the
trajectories stay in the tubes. Taking the relevant suprema proves
\eqref{eq:ua_local_bound}. The strict tolerance in
\eqref{eq:ua_local_tolerance} gives corrected error less than $\eta$.
No global residual regularity or globally defined policy is used.\hfill$\square$

\paragraph{Local sensitivity to companion samples.}
If $\|u-f(x)\|_2\le\rho a_f^{-K}$, the same induction for a designated
sample gives
\[
 \|Y_K(x,u;\xi)-Y_K(x,u;\xi')\|_2
 \le2a_f^K\|u-f(x)\|_2,
\]
provided both realized configurations have well-defined finite matrices
and satisfy the same lower bound on the damping along this sample's trajectory.
For policies specified only on full batch histories inside the tubes,
Proposition~\ref{prop:ua_local} ensures these conditions when all their
initial states satisfy the stated tolerance. If companions are arbitrary,
the step assumptions along the realized configurations must instead be
checked directly.

\begin{lemma}[Sufficient conditions for local residual growth]
\label{lem:ua_local_sufficient}
For the residual in~\eqref{eq:residual}, a bound
\[
 \|c_x(y)-c_x(f(x))\|_2\le L_f\|y-f(x)\|_2
 \quad\text{for }(x,y)\in Q_f(\rho)
\]
implies~\eqref{eq:ua_local_growth}. In particular, such a bound holds if
$c_x$ is differentiable in $y$ and $J_c(x,y)$ is jointly continuous on a
neighborhood of the compact tube $Q_f(\rho)$.
\end{lemma}
\begin{proof}
The first assertion follows from~\eqref{eq:ua_residual_comparison}.
For the second, compactness gives
\[
 L_f=\sup\limits_{(x,v)\in Q_f(\rho)}\|J_c(x,v)\|_2<\infty.
\]
For fixed $x$, the segment from $f(x)$ to $y$ stays in the same radius-$\rho$
ball. The mean-value integral therefore gives
\[
 c_x(y)-c_x(f(x))
 =\int_0^1J_c\bigl(x,f(x)+t(y-f(x))\bigr)(y-f(x))\,dt,
\]
and its norm is bounded by $L_f\|y-f(x)\|_2$.
\end{proof}

For the polar-voltage, power-balance AC-OPF formulation in
Section~\ref{sec:acopf}, the constraints are built from smooth power-flow
expressions, affine bounds, and squared apparent-power limits
\citep{zimmerman2019opfderivatives}. Their derivatives are continuous on
compact target tubes. If unsquared flow magnitudes are used, the
inequality $|\|v\|_2-\|w\|_2|\le\|v-w\|_2$ yields local Lipschitzness
instead; differentiability of the magnitude at zero is unnecessary for
residual growth. In either case, applying the correction still requires
finite matrices at every visited state, as specified in Appendix~\ref{app:ua_policies}.

The AC-OPF applicability statement concerns continuous feasible targets.
It does not assert continuity of an optimal solution map or exact
feasibility after finitely many corrections. The tolerance in
Proposition~\ref{prop:ua_local} may depend on $K$; the uniform bound of
Corollary~\ref{cor:ua_uniform} requires the additional global assumptions
stated there.

\section{Proof of Proposition~\ref{prop:screened_convergence}}
\label{app:screened_convergence}
Fix a sample and suppress its input and sample index. Since
$\nabla V(y_k)=J_{c,k}^\top r_k$ and $P_k^\top P_kr_k=r_k$, the screened
update can be written as
\[
 y_{k+1}=y_k-B_k\nabla V(y_k),\qquad
 B_k=(J_{c,k}^\top P_k^\top P_kJ_{c,k}+\varepsilon I)^{-1}.
\]
The global Lipschitz bound for the differentiable map $c$ gives
$\|J_{c,k}\|_2\le G$. Since $0\preceq P_k^\top P_k\preceq I$,
\[
 \|\tilde J_k\|_2\le G,\qquad
 \lambda_{\max}(\tilde J_k\tilde J_k^\top)\le G^2,\qquad
 \frac1{\varepsilon+G^2}I\preceq B_k\preceq\frac1\varepsilon I.
\]
In particular, the eigenvalues of $B_k$ lie in $(0,2/L)$. The concavity
of $t\mapsto t-Lt^2/2$ on this interval and the spectral bounds give
$B_k-(L/2)B_k^2\succeq\gamma(\varepsilon)I$.
The smoothness inequality therefore yields
\begin{align*}
 V(y_{k+1})
 &\le V(y_k)-\nabla V(y_k)^\top B_k\nabla V(y_k)
                  +\frac L2\|B_k\nabla V(y_k)\|_2^2\\
 &\le V(y_k)-\gamma(\varepsilon)\|\nabla V(y_k)\|_2^2
 \le\bigl(1-2\mu\gamma(\varepsilon)\bigr)V(y_k).
\end{align*}
We have $0<\gamma(\varepsilon)\le1/(2L)$, so $0<\mu<L$ gives
$q\in(0,1)$. Iteration proves the claim. The argument applies to every
realized sample trajectory and uses only the displayed uniform bounds;
dependence of $P_k$ on companion samples or previous iterates does not
alter the estimate.\hfill$\square$

\section{AC-OPF Details}\label{app:opf_details}
\subsection{Motivating application: Optimal Power Flow (OPF)}\label{sec:acopf}
Optimal Power Flow (OPF) describes an optimization problem grid operators must solve to determine power dispatch of their generators. Operators receive power demand at various buses, and must dispatch generators subject to capacity constraints, giving the following nonconvex optimization problem. 
\begin{align}
\min_{\substack{
    \{P_i^G,Q_i^G,V_i,\theta_i\}_{i\in\mathcal N}
}}
\quad &
\sum_{i\in\mathcal G}
\left(
    c_{2i}(P_i^G)^2 + c_{1i}P_i^G + c_{0i}
\right)
\label{eq:acopf_obj}
\\[0.5em]
\text{s.t.}\quad
&
P_i^G - P_i^D
=
\sum_{j\in\mathcal N_i} P_{ij}(V,\theta),
\qquad \forall i\in\mathcal N,
\label{eq:active_balance}
\\
&
Q_i^G - Q_i^D
=
\sum_{j\in\mathcal N_i} Q_{ij}(V,\theta),
\qquad \forall i\in\mathcal N,
\label{eq:reactive_balance}
\\
&
P_i^{G,\min}
\leq P_i^G
\leq P_i^{G,\max},
\qquad \forall i\in\mathcal G,
\label{eq:pg_limits}
\\
&
Q_i^{G,\min}
\leq Q_i^G
\leq Q_i^{G,\max},
\qquad \forall i\in\mathcal G,
\label{eq:qg_limits}
\\
&
V_i^{\min}
\leq V_i
\leq V_i^{\max},
\qquad \forall i\in\mathcal N,
\label{eq:voltage_limits}
\\
&
\theta_{\text{ref}}=0,
\label{eq:reference_bus}
\\
&
P_{ij}(V,\theta)^2+Q_{ij}(V,\theta)^2
\leq \left(\overline S_{ij}\right)^2,
\qquad \forall (i,j)\in\mathcal E.
\label{eq:thermal_limits}
\end{align}

Here, $\mathcal{N}$ represents the set of buses, $\mathcal{G}$ represents the set of generators, and $\mathcal{E}$ represents the set of transmission lines connecting individual buses. $P^G$ and $Q^G$ are the active and reactive power, respectively, generated at each generator, while $P^D$ and $Q^D$ are the active and reactive power, respectively, demanded at each bus. Active and reactive power generation is bounded at each generator with lower bounds $P_i^{G,\text{min}}$,$Q_i^{G,\text{min}}$ and upper bounds $P_i^{G,\text{max}}$,$Q_i^{G,\text{max}}$. $V$ is the voltage magnitude and $\theta$ is the voltage angle applied at each bus. $P_{ij}(V,\theta)$ and $Q_{ij}(V,\theta)$ are the active and reactive power transmitted between bus $i$ and $j$, determined by the voltages and line impedances, and limited by transmission capacity constraints represented by $\bar{S}_{ij}$.

Often, this problem is solved many times for many different instances (i.e., for many different demand forecasts). Furthermore, the number of buses can number in the hundreds or thousands, meaning a large-scale nonconvex optimization problem must be solved for each instance. The computational burden of such a scheme motivates the use of machine learning-based optimization surrogates. The idea is to map the problem instance (the demand at every bus) to the optimal decision variables (power generation and voltages at each bus) subject to the constraints ~\eqref{eq:active_balance}-\eqref{eq:thermal_limits}. These mappings can offer improved latency over classical solvers that must start from scratch at each instance. The ability for these surrogate solvers to enforce constraint satisfaction is therefore crucial for their practical implementation. This motivates the present work about efficient, scalable approaches for nonlinear, nonconvex constraint enforcement for neural networks. 

\subsection{AC-OPF implementation details}

The constraints are summarized in Table~\ref{tab:case118-constraints}. We always retain the power-balance~\eqref{eq:active_balance},~\eqref{eq:reactive_balance} and reference-angle equalities~\eqref{eq:reference_bus}, as well as generator~\eqref{eq:pg_limits}~\eqref{eq:qg_limits} and voltage-limit~\eqref{eq:voltage_limits} constraints, and screen only the thermal branch flow constraints~\eqref{eq:thermal_limits}. This is physically well-motivated because thermal limits are relevant only on congested transmission lines. Under normal operating conditions, power is distributed across many branches and most lines operate below their thermal ratings. Thus, only a small subset of branch-flow constraints is typically active or nearly active. A flow row is retained when $\frac{|S_{ij}(y)|}{\bar S_{ij}}\geq 1-\delta$, where
 $\delta\in[0,1]$, and $(\bar S_{ij})$ is the branch thermal rating. Here, $\delta=0$ retains only active or violated thermal constraints (a $J_r$-like treatment of the flow rows), while ($\delta=1$) retains every constraint and recovers the full constraint Jacobian ($J_c$).

\begin{table}[!t]
  \centering
  \caption{Constraints for the IEEE 118-bus experiment.}
  \label{tab:case118-constraints}
  \begin{tabular}{@{}llcr@{}}
    \toprule
    Constraint & Expression & Type & Count \\
    \midrule
    Active power balance
      & $P_i^G - P_i^D - 
\sum_{j\in\mathcal N_i} P_{ij}(V,\theta) = 0$
      & Eq. & 118 \\
    Reactive power balance
      & $Q_i^G - Q_i^D -
\sum_{j\in\mathcal N_i} Q_{ij}(V,\theta)=0$
      & Eq. & 118 \\
    Slack angle reference
      & $\theta_{\mathrm{ref}} = 0$
      & Eq. & 1 \\
    \midrule
    Active generation limits
      & $P_i^{G,\min} \leq P_i^G \leq P_i^{G,\max}$
      & Ineq.$^\dagger$ & 54 \\
    Reactive generation limits
      & $Q_i^{G,\min} \leq Q_i^G\leq Q_i^{G,\max}$
      & Ineq.$^\dagger$ & 54 \\
    Voltage magnitude limits
      & $V_i^{\min}\leq V_i\leq V_i^{\max}$
      & Ineq.$^\dagger$ & 118 \\
    Branch flow limits (from end)
      & $|S^f_\ell(v,\theta)| \le \overline{s}_\ell$
      & Ineq. & 186 \\
    Branch flow limits (to end)
      & $|S^t_\ell(v,\theta)| \le \overline{s}_\ell$
      & Ineq. & 186 \\
    \bottomrule
  \end{tabular}

  \vspace{2pt}
  {\footnotesize $^\dagger$Two-sided bound, counted once per variable
  (452 one-sided inequalities).}
\end{table}

\end{document}